# Evaluating Large Language Models for Health-Related Text Classification Tasks with Public Social Media Data

Yuting Guo[1], Anthony Ovadje[2], Mohammed Ali Al-Garadi[3], Abeed Sarker[1]

[1]Department of Biomedical Informatics, Emory University, Atlanta, GA, USA.
[2]Department of Biomedical Engineering, Georgia Institute of Technology, Atlanta, GA, USA
[3]Vanderbilt University Medical Center, Vanderbilt University, Nashville, TN, USA
yuting.guo@emory.edu

**Abstract.** Large language models (LLMs) have demonstrated remarkable success in NLP tasks. However, there is a paucity of studies that attempt to evaluate their performances on social media-based health-related natural language processing tasks, which have traditionally been difficult to achieve high scores in. We benchmarked one supervised classic machine learning model based on Support Vector Machines (SVMs), three supervised pretrained language models (PLMs) based on RoBERTa, BERTweet, and SocBERT, and two LLM based classifiers (GPT3.5 and GPT4), across 6 text classification tasks. We developed three approaches for leveraging LLMs for text classification: employing LLMs as zero-shot classifiers, using LLMs as annotators to annotate training data for supervised classifiers, and utilizing LLMs with few-shot examples for augmentation of manually annotated data. Our comprehensive experiments demonstrate that employing data augmentation using LLMs (GPT-4) with relatively small human-annotated data to train lightweight supervised classification models achieves superior results compared to training with human-annotated data alone. Supervised learners also outperform GPT-4 and GPT-3.5 in zero-shot settings. By leveraging this data augmentation strategy, we can harness the power of LLMs to develop smaller, more effective domain-specific NLP models. LLM-annotated data without human guidance for training lightweight supervised classification models is an ineffective strategy. However, LLM, as a zero-shot classifier, shows promise in excluding false negatives and potentially reducing the human effort required for data annotation. Future investigations are imperative to explore optimal training data sizes and the optimal amounts of augmented data.

**Keywords:** Text classification, large language models, natural language processing.

## 1 Introduction

Social media platforms serve as a valuable medium for patients to share and discuss their health-related information, encompassing a broad spectrum of topics. To derive knowledge about these topics, researchers have employed natural language processing (NLP) technologies, often employing text classification methods. Supervised classifi-cation of social media data is particularly challenging, relative to texts from other sources, due to the inherent noise. The linguistic characteristics of the text can vary significantly depending on the originating social media platform. For instance, Twitter (rebranded as X) posts commonly feature hashtags and emojis, whereas this is not typical for posts on Reddit. Another challenge in text classification for health-related tasks involving social media data lies in the efficiency of data collection. The traditional data collection pipeline for social media comprises two key steps: 1) filtering out unrelated posts using a keyword list, and 2) manually annotating the selected posts. Given that most social media content is typically unrelated to the research topic of interest, the resultant dataset often exhibits a markedly reduced size, or the class distribution tends to be highly unbalanced.

Pretrained language models (PLMs) such as BERT [1] and RoBERTa [2] have demonstrated remarkable success in a wide range of NLP tasks. Encouraged by the success of transformer-based PLMs, many studies have focused on adapting them to text classification tasks involving social media data. Nguyen et al. [3] proposed BERTweet by pretraining a transformer-based model from scratch on a large set of English posts from Twitter. Guo et al. [4] developed SocBERT which was pretrained on English posts from Twitter and Reddit. Qudar et al. [5] developed TweetBERT by continuing pretraining the language model of BERT. Additionally, research efforts have been taken to organize shared tasks or competitions for text classification for health-related topics in social media. For example, the Social Media Mining for Health (SMM4H) shared tasks have covered a wide range of health-related topics over the years including pharmacovigilance, toxicovigilance, and epidemiology, and involved data from different social media platforms such as Twitter and Reddit [6–12]. Additionally, the Sixth Workshop on Noisy User-generated Text (W-NUT 2020) proposed a text classification task aiming to identify informative COVID-19 English posts from Twitter [13].

Although the PLMs have achieved success, the model performance is highly dependent on the quality and quantity of annotated data of the downstream tasks. Researchers have made attempts to reduce the need for annotated data by pretraining larger language models (LLMs) with a generative model architecture such as GPT3 [14]. In 2023, a chatbot named ChatGPT powered by an LLM named GPT3.5 with 175B parameters and pretrained on a large corpus of text data from the Internet, achieved great success on various question-answering NLP tasks. Kung et al. [15] showed that ChatGPT performed at or near the passing threshold for the United States Medical Licensing Exam without requiring supervision with human-annotated data. Chen et al. [16] examined the performance of ChatGPT on various neurological exam grading scales, where ChatGPT demonstrated ability in evaluating neuroexams using established assessment scales. Similarly, Dehghani et al. [17] evaluated the performance of ChatGPT on a radiology board-style examination, and ChatGPT correctly answered 69% of questions. Despite their primary purpose as generative models for text generation, LLMs have been effectively utilized in text classification endeavors. Some studies have directly tasked LLMs with predicting classifications in a zero-shot setting, a machine learning approach that functions without prior training data [18–23]. Additionally, research has delved into investigating techniques for integrating LLMs into text classification tasks, such as leveraging them for data augmentation [24–26] and using LLMs to obtain external knowledge [27,28].

Inspired by the success of LLMs, we explore strategies for leveraging them for social media based health-related text classification. Potentially, via effective zero-shot classification, LLMs may significantly diminish the time and cost associated with data annotation. In this work, we sought to explore this potential by leveraging LLMs in three different settings: employing LLMs as zero-shot classifiers, using LLMs as annotators to annotate training data for supervised classifiers, and utilizing LLMs for data augmentation. Our contributions are outlined below:

- We developed and compared three approaches for integrating LLMs into text classification.
- We conducted a comprehensive benchmarking exercise, evaluating one supervised classic machine learning model based on Support Vector Machines (SVMs) [29], three supervised PLMs (RoBERTa [2], BERTweet [3], and SocBERT [4]), and two zero-shot classifiers based on GPT3.5 [30] and GPT4 [31], across 6 text classification tasks.
- Our findings demonstrate that the most optimal strategy is to leverage in-context trained LLMs for augmenting human-annotated data. Nevertheless, future investigations are needed to explore the determination of appropriate training data size and the optimal volume of augmented data.

**Table 1.** The number of positive and negative classes and data sizes for 6 classification tasks.

| Task | Positive | Negative | Total |
|---|---|---|---|
| Self-report depression | 625 | 477 | 1102 |
| Self-report COPD | 401 | 373 | 774 |
| Self-report breast cancer | 1283 | 3736 | 5019 |
| Change in medications regimen | 656 | 6814 | 7470 |
| Self-report adverse pregnancy outcomes | 2922 | 3565 | 6487 |
| Self-report potential cases of COVID-19 | 1148 | 6033 | 7181 |

# 2 Methods

## 2.1 Data Collection

In this work, we involved 6 text classification tasks covering diverse healthcare topics with data from Twitter. Among these tasks, 4 tasks—classification of self-report breast cancer, classification of change in medications regimen, classification of self-report of adverse pregnancy outcomes, and classification of self-report potential cases of COVID-19, are from Social Media Mining for Health Applications (SMM4H) shared tasks [11], and 2 tasks—classification of self-report depression and self-report Chronic Obstructive Pulmonary Disease (COPD) are private data sets collected and annotated by our team. All of the tasks are binary classification and the evaluation metrics are precision, recall, and $F_1$ score for the positive class. The data statistics of the tasks are shown in Table 1.

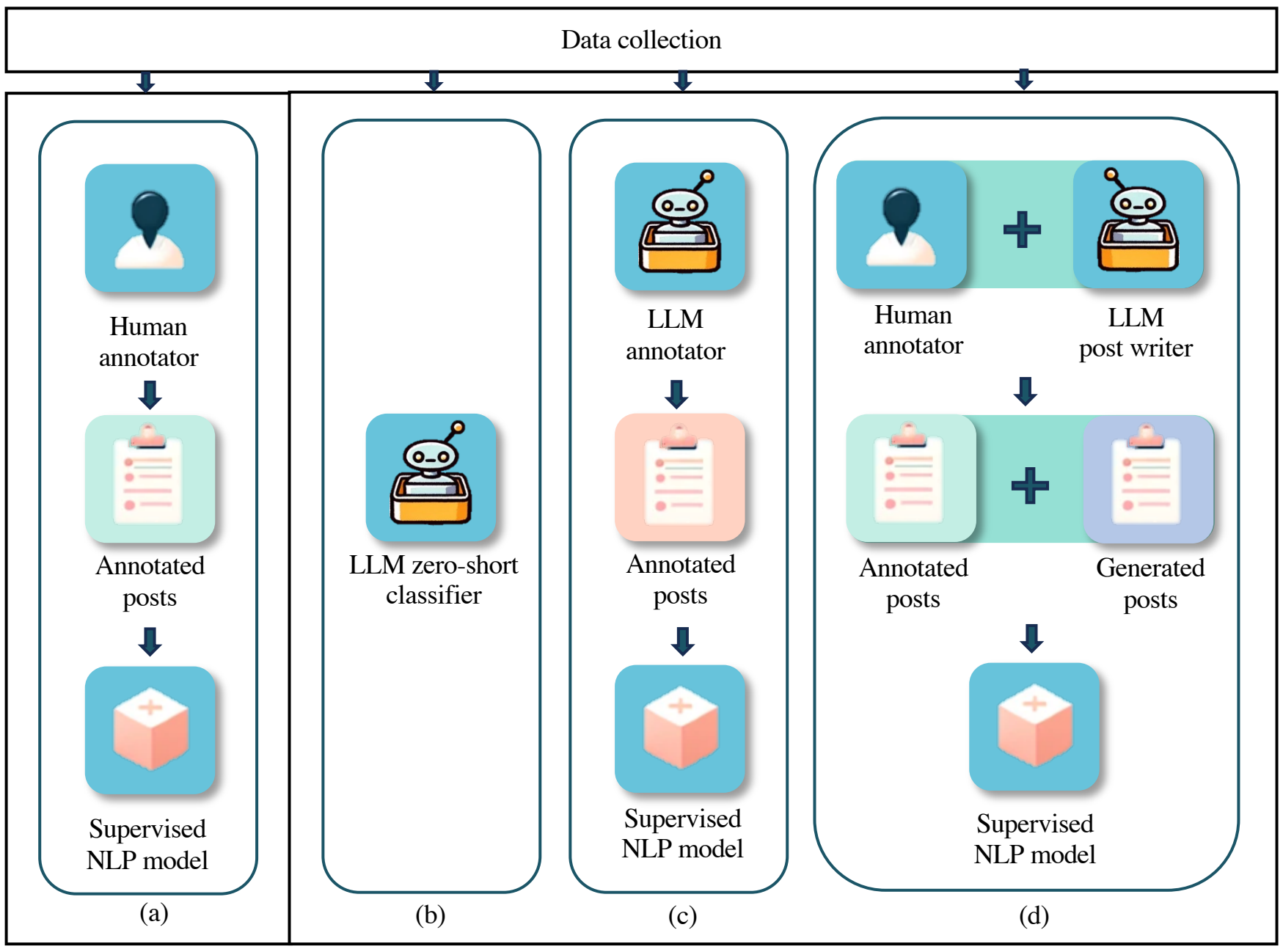

**Fig. 1.** The framework for comparing four supervised text classification strategies for benchmarking: (a) standard supervised classification approach with manually annotated data; (b) employing the LLM as a zero-shot classifier; (c) using the LLM as an annotator for data that can be used by supervised classification approaches; and (d) utilizing LLMs for data augmentation for traditional supervised classification models.

## 2.2 Classification Models

The overall classifier benchmarking framework of this study is shown in Fig. 1. We developed the supervised NLP models including one supervised classic machine learning model based on SVMs, and three supervised PLMs (RoBERTa, BERTweet, and SocBERT). We refer to these as *traditional* supervised classification approaches. We also developed three approaches for integrating LLMs into text classification.

These methods include employing the LLM as a zero-shot classifier, using the LLM as an annotator to annotate training data, and utilizing the LLM for data augmentation. We opted for GPT3.5 and GPT4 as the LLM in the experiments due to their widespread popularity and ready accessibility within the research community.

For the supervised NLP models, we divided the data using stratified 80-20 random splits to ensure that the class distributions of the training and test set remain the same. The training set was used for training and optimization of models, and the test set was used for evaluating model performances. The same data splits were used for developing and evaluating the LLM based approaches. The evaluation metrics were the precision, recall, and $F_1$ score over the positive class on the test set, where $F_1$ score was chosen as the primary metric for comparison because it ensures that neither precision (positive predictive value) nor recall (sensitivity) is optimized at the expense of the other. For all models, 5-fold cross-validation was performed to measure the classification performance, and the mean and standard deviation of the evaluation metrics were computed.

**SVM**

SVMs are often used when dealing with large feature spaces, making them a popular choice for text classification, including in this study. To represent the text notes as features, we used Term Frequency - Inverse Document Frequency (TF-IDF) for n-grams (sequences of n-words). In this case, 1-, 2-, 3-, and 4-grams were used. In the training process, grid search was employed to find the best values for two important hyperparameters: the kernel function (K) which could be either "linear" or "RBF", and the C value [2, 4, 6, 8, 10]. Furthermore, a class weighting strategy was implemented that automatically adjusts weights based on the inverse proportion of class frequencies in the input data, meaning the majority class receives a lower weight compared to the minority class during training, which can address data imbalance issues.

**PLM based Classification**

The model architecture for PLM based classification contains a PLM encoder that converts the input text into a vector representation. Unlike the machine learning models that extract text-based features or generate n-grams, a PLM based classification model splits a document into word pieces or tokens. Each token is then encoded into a vector, and these vectors are combined to form a vector representation of the document. The vector representation is then fed into a classification layer and an output layer with a softmax function. This produces a vector of equivalent size to the number of classes, from which the class with the highest value in its corresponding dimension is selected. In this study, we used three PLMs—RoBERTa, BERTweet, and SocBERT. These models have the same model architecture but were pretrained on different data sources. RoBERTa was pretrained on generic web text, BERTweet was pretrained on the posts from Twitter only, and SocBERT was pretrained on posts from Twitter and Reddit. Hyperparameters for the three models and other relevant details are listed in Table S1 in the supplementary material.

## Leveraging LLMs

*LLM Zero-shot Classifier/Annotator*

In the context of using the LLM as an annotator, we asked the LLM to give the label of one sample, and the LLM responded in the way we requested. The study of how to ask the model to do this work is called prompt engineering. Because the model performance is highly dependent on the prompt, we used the same template for all classification tasks and only changed the task-specific keywords when running a task. Additionally, we asked the model to reply either 1 or 0 which indicated positive or negative class. Being generative models, the responses sometimes did not follow the instructions provided, and we considered those to be 0. The prompt template was as follows:

"*You are a [X] system based on raw tweet data. The system should analyze the provided tweet and predict whether the user is [X] or not. Given a tweet as input, the system should output a 1 if the user is [X], and 0 otherwise. If a text response is generated, reanalyze the input until a 1 or 0 is generated.*"

where *[X]* denotes the task-specific words. The detailed prompts for all classification tasks are in supplementary Table S2. After using the LLM as an annotator to annotate training data, we trained SVM and RoBERTa on the training data with the model-annotated labels. We evaluated on the test data with the human-annotated labels.

When utilizing the LLM as a zero-shot classification model, we leveraged the predictions generated in the experiment where the LLM served as an annotator. The classification performance evaluation was performed on the same test set, utilizing labels annotated by humans.

*Write* ***5*** *tweets close to the tweet [text]. The output should follow this format:*
*tweet 1:[tweet]*
*tweet 2:[tweet]*
*tweet 3:[tweet]*
*tweet 4:[tweet]*
*tweet 5:[tweet]*

**Fig. 2.** The prompt for data augmentation where *[text]* was the placeholder for the original post, and *[tweet]* was the placeholder for the post generated by the LLM.

*Using LLMs for Data Augmentation*

In the realm of machine learning, it is a well-established fact that increasing the volume of data can significantly boost a model's performance. To harness this principle, we employed a LLM to augment our training data for classification tasks. Our approach involved instructing the LLM to generate posts closely resembling each entry in our original training set. The specified prompt was structured as Fig. 2. Subse-quently, we assigned identical labels to the generated posts, creating a new training set that encompassed both the original posts and their LLM-generated counterparts. This expansion resulted in a training set five times larger than its initial size. Table S3 showcases one original post and its corresponding five generated posts. We trained the supervised NLP models on this artificially augmented dataset. Subsequent evaluation was conducted on the test data, annotated by human experts. To investigate the impact of data size on model performance, we conducted a comparative analysis of training data comprising 1, 2, 3, 4, and 5 generated posts for each original post. Additionally, we explored the efficacy of data augmentation by varying the percentage of augmented training data (10%, 20%, ..., 100%). This inquiry also sought to ascertain whether leveraging LLM for data augmentation could alleviate the manual effort required for data annotation.

We employed GPT3.5 and GPT4 in conducting data augmentation experiments for self-report depression and self-report COPD tasks. The rationale behind selecting these tasks was rooted in their relatively small data sizes. The cost associated with API usage for data augmentation surpassed that of classification, prompting our decision.

# 3 Results

## 3.1 Classification Results

Fig. 3 displays the classification outcomes obtained through three approaches: employing human-annotated data as training data, utilizing LLM-annotated data as training data, and utilizing the LLM as a zero-shot classifier. Across all tasks, the supervised PLM based models, namely RoBERTa, BERTweet, and SocBERT, which were trained on data annotated by humans, demonstrated superior performance compared to their counterparts that were trained on LLM-annotated data. The averaged $F_1$ score differences for RoBERTa, BERTweet, and SocBERT trained on human annotated data compared to their counterparts trained on GPT3.5-annotated data were 0.24 (std: ±0.10), 0.25 (±0.11), and 0.23 (±0.11); compared to their counterparts trained on GPT4-annotated data, the differences were 0.16 (±0.07), 0.16 (±0.08), and 0.14 (±0.08), respectively. Across all tasks, the averaged $F_1$ score difference between the top performing model and the GPT3.5 zero-shot classifier was 0.26 (±0.11), and for the GPT4 zero-shot classifier, it was 0.15 (±0.08). However, an intriguing contrast emerged when examining the supervised SVM models. Those trained on human-annotated data exhibited inferior performance compared to their counterparts trained on GPT3.5-annotated data and GPT4-annotated data in two specific tasks: self-reported depression and self-reported COPD. Furthermore, the SVM model trained on GPT4-annotated data outperformed that trained on human-annotated data in the task of self-reported potential cases of COVID-19. Looking at the LLM zero-shot classifiers, the GPT3.5 zero-shot classifier outperformed SVM for a single task (self-report depression), while the GPT4 zero-shot classifier outperformed SVM in 5 out of 6 tasks (except for self-report COPD). In contrast, the PLM based models trained on human-annotated data consistently outperformed the LLM zero-shot classifiers, while the GPT4 zero-shot classifier achieved higher *recall* than the models trained on human-annotated data in 5 out of 6 tasks (except for self-report adverse pregnancy outcomes). We also observed the differences between the two LLMs. Across all tasks, the GPT4 zero-shot classifier consistently delivered a higher $F_1$ score compared to the GPT3.5 zero-shot classifier. Additionally, the supervised NLP models trained on GPT4 annotated data exhibited superior performance when compared to their counterparts trained on GPT3.5 annotated data across most tasks. These findings suggest that predictions generated by GPT4 tend to be more accurate than those from GPT3.5, which aligned with expectations given that GPT4 can be considered a more advanced version of GPT3.5.

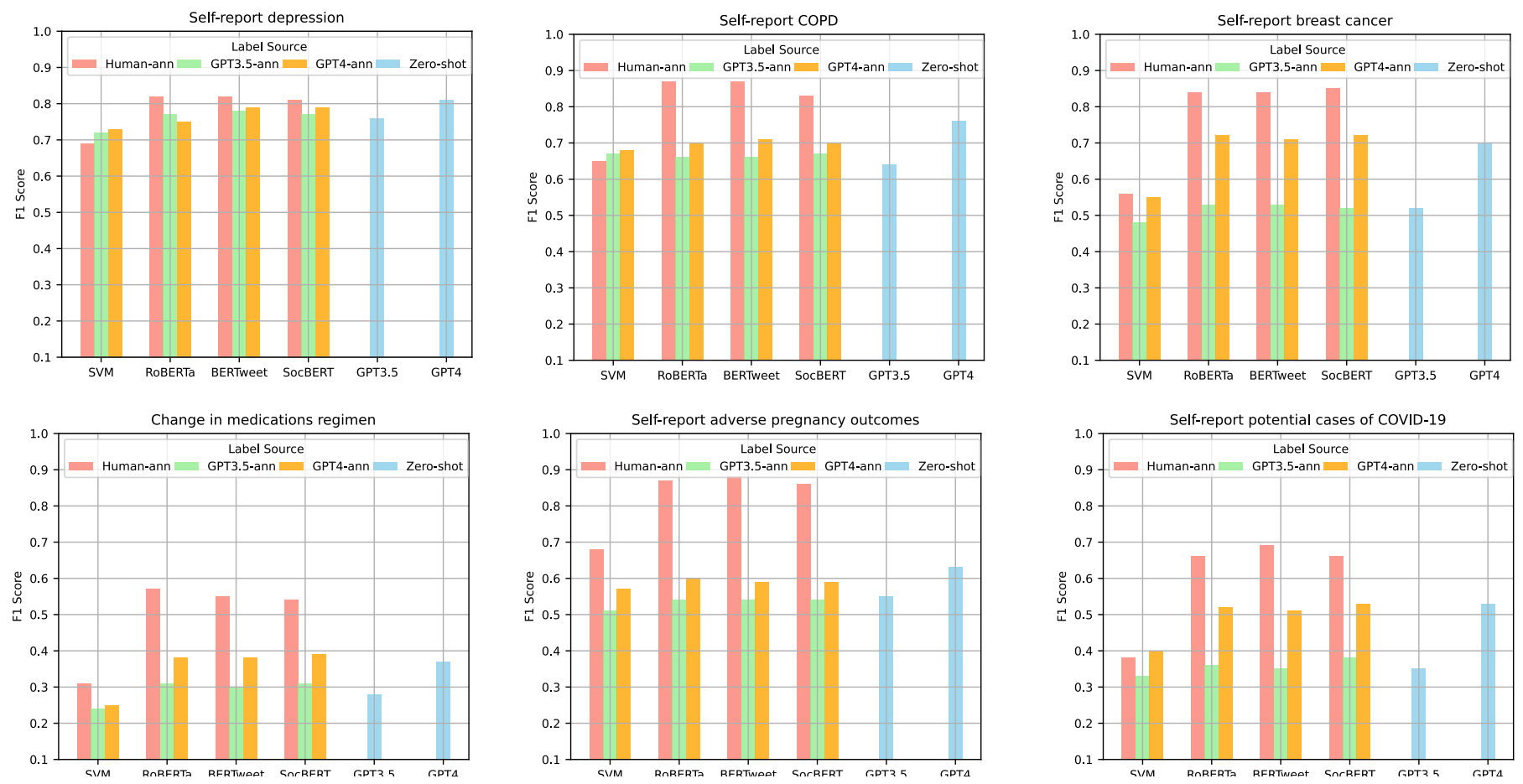

**Fig. 3.** The classification results of employing human-annotated data as training data (Human-ann), utilizing LLM-annotated data as training data (GPT3.5-ann and GPT4-ann), and utilizing the LLM as a zero-shot classifier (GPT3.5 zero-shot and GPT4 zero-shot). The detailed precision, recall, and $F_1$ scores are listed in supplementary Table S4.

## 3.2 Data augmentation results

We opted for RoBERTa as the supervised NLP model for examining the efficacy of data augmentation. This choice was motivated by the superior performance of PLM based models over SVM. Moreover, among the three PLMs, RoBERTa has been subjected to more extensive research. Fig. 4 visually represents the outcomes of RoBERTa models trained on GPT3.5 augmented data, GPT4 augmented data, and the human annotated data for self-report depression and self-report COPD classification, respectively. Broadly, the classification performance trend across both tasks remained consistent. In comparison to models trained solely on human-annotated data, those trained on GPT4 augmented data demonstrated superior or comparable performance. However, models trained on GPT3.5 augmented data exhibited worse or comparable results. The optimal performance, particularly with GPT4 augmented data, was achieved when utilizing 60% of the training data for self-report of depression classifi-cation and 100% of the training data for self-report COPD repression. For self-report of depression, the optimal number of generated posts for its initial post was 5, while for self-report COPD, it was 2. We noted that the effectiveness of results did not necessarily correlate with the quantity of generated posts. The impact of increasing the number of generated posts can vary depending on the specific task being addressed.

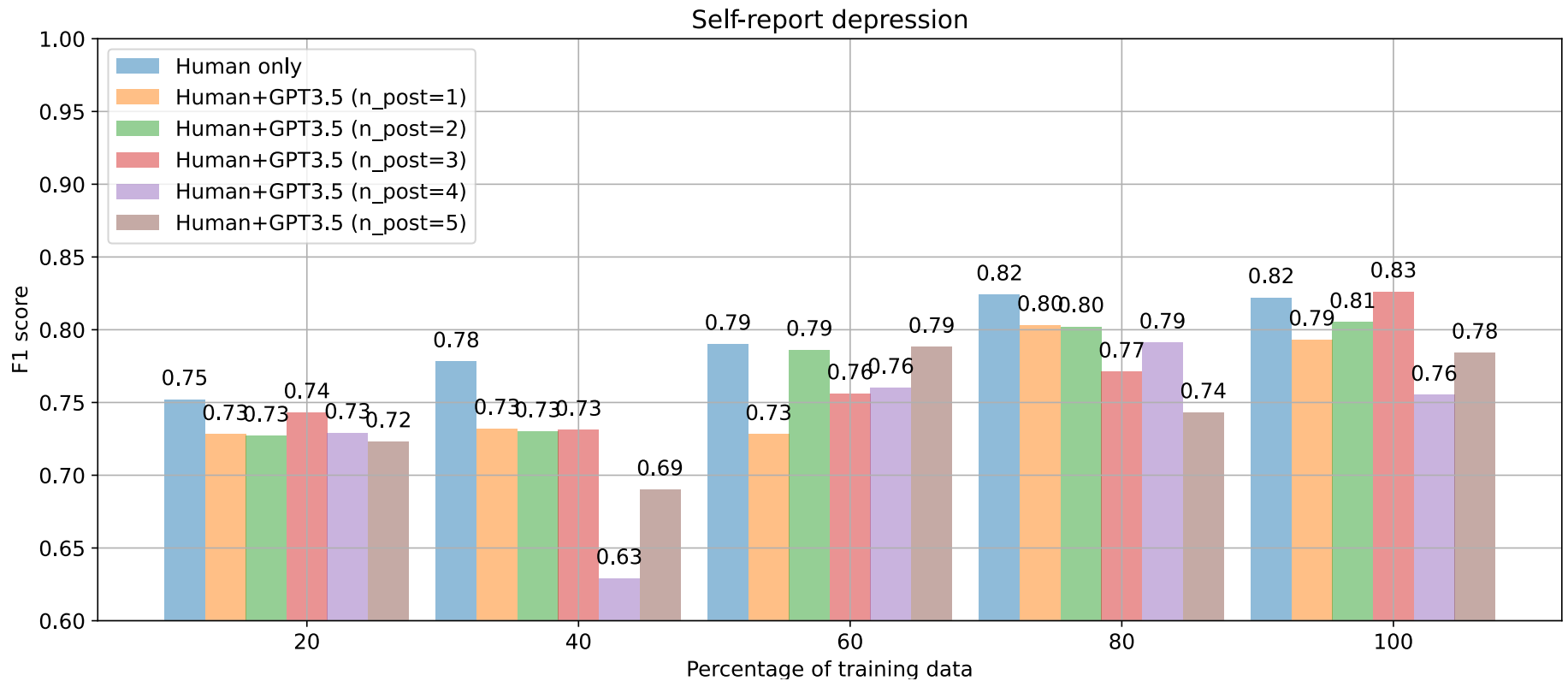

(a) The results of RoBERTa trained on GPT3.5 augmented data and human annotated data for self-report depression classification.

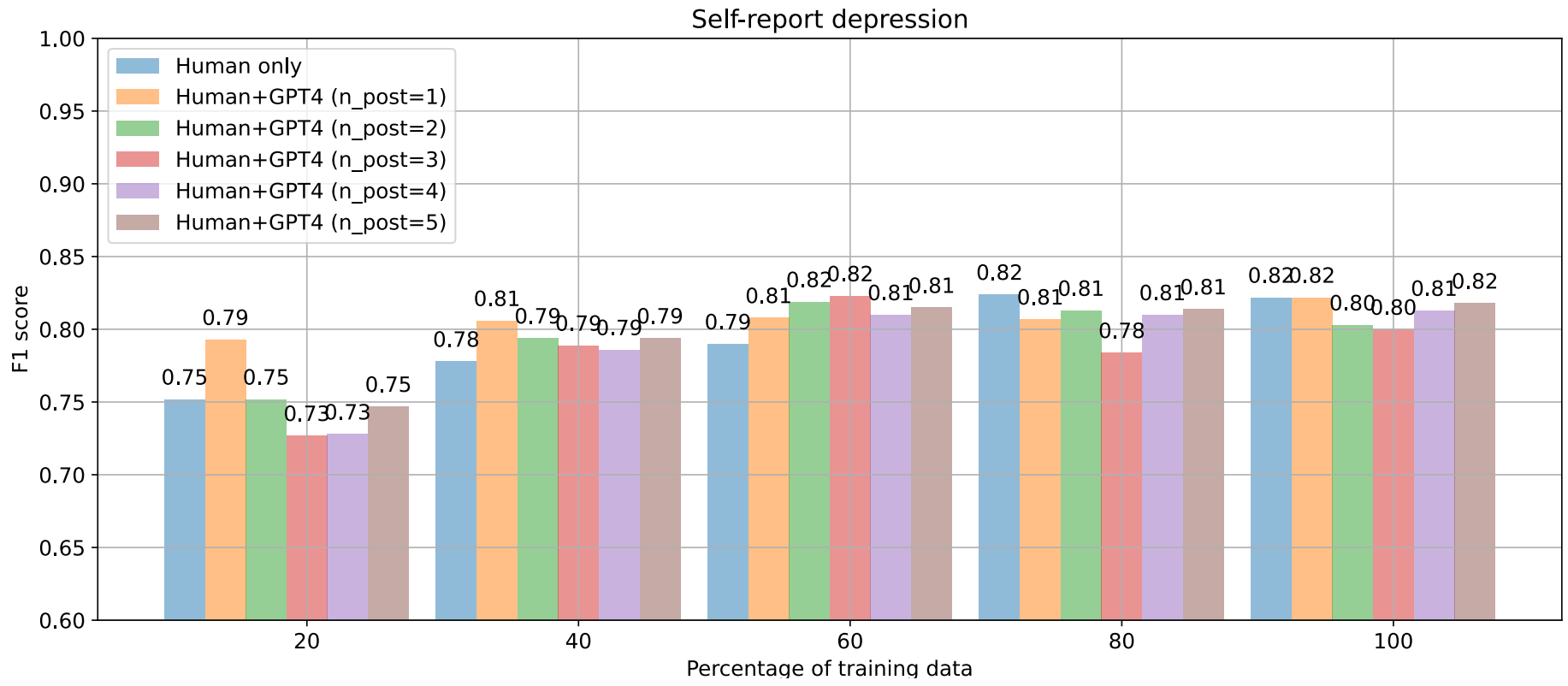

(b) The results of RoBERTa trained on GPT4 augmented data and human annotated data for self-report depression classification.

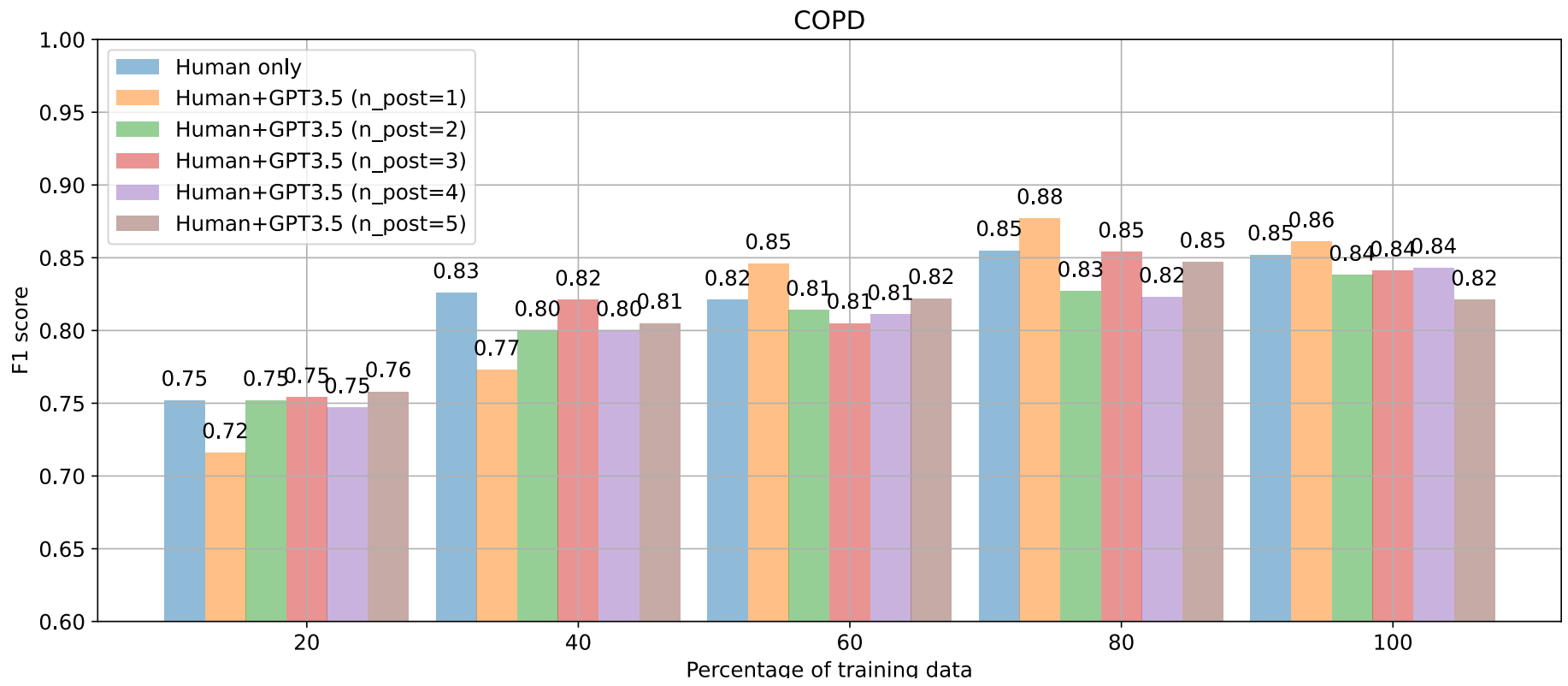

(c) The results of RoBERTa trained on GPT3.5 augmented data and human annotated data for self-report depression classification.

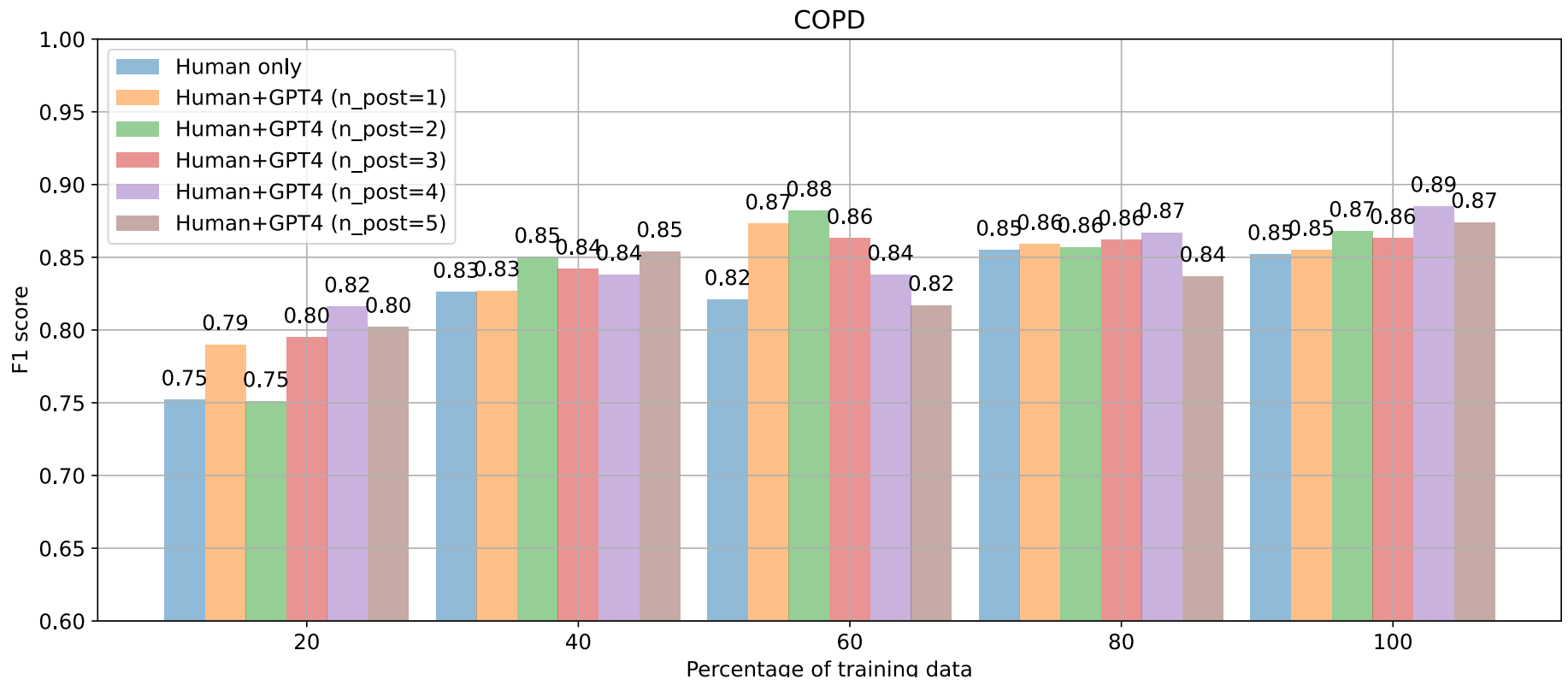

(d) The results of RoBERTa trained on GPT4 augmented data and human annotated data for self-report depression classification.

**Fig. 4.** The $F_1$ scores for the RoBERTa models trained on GPT3.5 augmented data, GPT4 augmented data, and the human annotated data alone for self-report depression and self-report COPD, respectively.

## 4 Discussion

We conducted a comprehensive comparison of classification performances by employing human-annotated data, LLM-annotated data as training data, and utilizing the LLM as a zero-shot classifier. The outcomes revealed that using LLM-annotated data only for training supervised classification models was ineffective. However, employing the LLM as a zero-shot classifier exhibited the potential to outperform traditional SVM models and achieved a higher recall than the advanced transformer-based model

RoBERTa. This suggests that the LLM zero-shot classifier may in some cases serve as a valuable preprocessing module to exclude false negatives before leveraging supervised classification models. This can be particularly useful for extremely imbalanced datasets. In the context of the self-report depression classification task, we observed a marginal performance gap between the GPT4 zero-shot classifier and the PLM based models trained on human-annotated data. This implies the promising prospect of using the LLM zero-shot classifiers to reduce the human effort required for data annotation.

We also obtained interesting results from our data augmentation experiments. Our results indicated that utilizing GPT3.5 for data augmentation could potentially harm model performance. This underscores the importance of the 'LM's capability to generate high-quality posts for effective data augmentation. In contrast, data augmentation with GPT4 demonstrated improved model performances, showcasing the potential of LLMs in reducing the need for extensive training data. Furthermore, we found that the optimal number of augmented data and the ideal ratio of human annotated data and LLM augmented data can be task-specific, emphasizing that increased augmented data may not lead to enhanced model performance. It remains an underexplored area and requires future investigation. We recommend researchers explore this task-specificity through grid searches to identify the most appropriate number of augmented data when applying data augmentation to a specific task. Overall, this study suggests, that the best performances for text classification may be obtained by performing some manual annotations followed by data augmentation via LLMs. Note also that the LLMs we used were neither fine-tuned for social media data nor medical texts. Customizing LLMs before data augmentation may further improve performance.

## 5 Limitation

Limitations of our study include the limited exploration of optimal prompts for enhancing LLM-based model performance, as our primary focus was on devising effective strategies for LLM integration into text classification. Future research should delve deeper into prompt engineering to ascertain its potential impact. Moreover, due to the prohibitive costs associated with utilizing proprietary models such as GPT3.5 and GPT4, we restricted our benchmarking and data augmentation experiments to a modest number of classification tasks. Consequently, our findings may not fully encapsulate the diverse range of potential applications. However, with the proliferation of open-source LLMs, future studies can extend our experiments across a wider array of datasets at a lower cost, thereby bolstering the generalizability of our conclusions.

## 6 Conclusions

In this study, we undertook a comprehensive examination of classification performances, utilizing human-annotated data, employing the LLM as a zero-shot classifier, using the LLM as an annotator to annotate training data, and utilizing the LLM for data augmentation. Our experiments demonstrate that combining data augmentation using LLMs (GPT-4) with human-annotated data to train lightweight supervised classification models achieves superior results compared to training with human-annotated data alone as well as outperforms zero-shot learning using GPT-4 and GPT-3.5 By leveraging this strategy, we can harness the power of LLMs to develop smaller, more effective domain-specific NLP models. The results indicate that using LLM-annotated data without human-annotated data for training lightweight supervised classification models is an ineffective strategy. However, LLM, as a zero-shot classifier, shows promise in excluding false negatives and potentially reducing the human effort required for data annotation. Future investigations are imperative to explore optimal training data sizes and the optimal amounts of augmented data.

## 7 Disclosures

The authors have no conflicts to declare.

## References

1. Devlin J. BERT: Pre-training of Deep Bidirectional Transformers for Language Understanding (Bidirectional Encoder Representations from Transformers). Bert-Ppt [Internet]. 2018; Available from: https://nlp.stanford.edu/seminar/details/jdevlin.pdf?utm_campaign=NLP News&utm_medium=email&utm_source=Revue newsletter

2. Liu Y, Ott M, Goyal N, Du J, Joshi M, Chen D, et al. RoBERTa: A robustly optimized BERT pretraining approach. arXiv. 2019;

3. Nguyen DQ, Vu T, Tuan Nguyen A, Nguyen AT. BERTweet: A pre-trained language model for english tweets. arXiv. Association for Computational Linguistics (ACL); 2020;9–14.

4. Guo Y, Sarker A. SocBERT: A Pretrained Model for Social Media Text. 2023 Work Insights from Negat Results NLP. 2023. p. 45–52.

5. Qudar MMA, Mago V. TweetBERT: A Pretrained Language Representation Model for Twitter Text Analysis. 2020;1–12. Available from: http://arxiv.org/abs/2010.11091

6. Sarker A, Nikfarjam A, Gonzalez-Hernandez G. Social Media Mining Shared Task Workshop. In: Altman RB, Dunker AK, Hunter L, Klein TE, Ritchie MD, editors. Biocomput 2016 Proc Pacific Symp Kohala Coast, Hawaii, USA, January 4-8, 2016 [Internet]. 2016. p. 581–92. Available from: http://psb.stanford.edu/psb-online/proceedings/psb16/sarker.pdf

7. Sarker A, Gonzalez-Hernandez G. Overview of the second social media mining for health (SMM4H) shared tasks at AMIA 2017. Training. 2017;1:1239.

8. Weissenbacher D, Sarker A, Paul MJ, Gonzalez-Hernandez G. Overview of the Third Social Media Mining for Health ({SMM}4{H}) Shared Tasks at {EMNLP} 2018. In: Gonzalez-Hernandez G, Weissenbacher D, Sarker A, Paul M, editors. Proc 2018 {EMNLP} Work {SMM}4{H} 3rd Soc Media Min Heal Appl Work {\&} Shar Task [Internet]. Brussels, Belgium: Association for Computational Linguistics; 2018. p. 13–6. Available from: https://aclanthology.org/W18-5904

9. Weissenbacher D, Sarker A, Magge A, Daughton A, O'Connor K, Paul MJ, et al. Overview of the Fourth Social Media Mining for Health ({SMM}4{H}) Shared Tasks at {ACL} 2019. In: Weissenbacher D, Gonzalez-Hernandez G, editors. Proc Fourth Soc Media Min Heal Appl Work {\&} Shar Task [Internet]. Florence, Italy: Association for Computational Linguistics; 2019. p. 21–30. Available from: https://aclanthology.org/W19-3203

10. Klein A, Alimova I, Flores I, Magge A, Miftahutdinov Z, Minard A-L, et al. Overview of the Fifth Social Media Mining for Health Applications ({\#}{SMM}4{H}) Shared Tasks at {COLING} 2020. In: Gonzalez-Hernandez G, Klein AZ, Flores I, Weissenbacher D, Magge A, O'Connor K, et al., editors. Proc Fifth Soc Media Min Heal Appl Work {\&} Shar Task [Internet]. Barcelona, Spain (Online): Association for Computational Linguistics; 2020. p. 27–36. Available from: https://aclanthology.org/2020.smm4h-1.4

11. Magge A, Klein AZ, Miranda-Escalada A, Al-Garadi MA, Alimova I, Miftahutdinov Z, et al. Overview of the Sixth Social Media Mining for Health

Applications (#SMM4H) Shared Tasks at NAACL 2021. 2021.

12. Weissenbacher D, Banda J, Davydova V, Estrada Zavala D, Gasco Sánchez L, Ge Y, et al. Overview of the Seventh Social Media Mining for Health Applications SMM4H Shared Tasks at COLING 2022. Proc Seventh Work Soc Media Min Heal Appl Work {\&} Shar Task [Internet]. Gyeongju, Republic of Korea: Association for Computational Linguistics; 2022. p. 221–41. Available from: https://aclanthology.org/2022.smm4h-1.54

13. Nguyen DQ, Vu T, Rahimi A, Dao MH, Nguyen LT, Doan L. WNUT-2020 Task 2: Identification of Informative COVID-19 English Tweets. Online. Association for Computational Linguistics (ACL); 2020. p. 314–8.

14. Brown TB, Mann B, Ryder N, Subbiah M, Kaplan J, Dhariwal P, et al. Language models are few-shot learners. Adv Neural Inf Process Syst. 2020;2020-Decem.

15. Kung TH, Cheatham M, Medenilla A, Sillos C, Leon L De, Madriaga M, et al. Performance of ChatGPT on USMLE : Potential for AI-Assisted Medical Education Using Large Language Models. 2022;3786:1–25.

16. Chen TC, Kaminski E, Koduri L, Singer A, Singer J, Couldwell M, et al. Chat GPT as a Neuro-Score Calculator: Analysis of a Large Language Model's Performance on Various Neurological Exam Grading Scales. World Neurosurg. 2023;179:e342–7.

17. Dehghani M, Djolonga J, Mustafa B, Padlewski P, Heek J, Gilmer J, et al. Scaling Vision Transformers to 22 Billion Parameters. Proc Mach Learn Res. 2023;202:7480–512.

18. Elhafsi A, Sinha R, Agia C, Schmerling E, Nesnas IAD, Pavone M. Semantic anomaly detection with large language models. Auton Robots. Springer; 2023;47:1035–55.

19. Krusche M, Callhoff J, Knitza J, Ruffer N. Diagnostic accuracy of a large language model in rheumatology: comparison of physician and ChatGPT4. Rheumatol Int. Springer; 2023;1–4.

20. Nazary F, Deldjoo Y, Di Noia T. ChatGPT-HealthPrompt. Harnessing the Power of XAI in Prompt-Based Healthcare Decision Support using ChatGPT. Commun Comput Inf Sci. 2024;1947:382–97.

21. Nedilko A. Generative pretrained transformers for emotion detection in a code-switching setting. Proc 13th Work Comput Approaches to Subj Sentim \& Soc Media Anal. 2023. p. 616–20.

22. Qiu Y, Jin Y. ChatGPT and finetuned BERT: A comparative study for developing intelligent design support systems. Intell Syst with Appl. Elsevier; 2024;21:200308.

23. Nicula B, Dascalu M, Arner T, Balyan R, McNamara DS. Automated Assessment of Comprehension Strategies from Self-Explanations Using LLMs. Information. MDPI; 2023;14:567.

24. Wang D, Ma W, Cai Y, Tu D. A general nonparametric classification method for multiple strategies in cognitive diagnostic assessment. Behav Res Methods. Springer; 2023;1–13.

25. Woźny M, Lango M. Alexander knox at semeval-2023 task 5: The comparison

of prompting and standard fine-tuning techniques for selecting the type of spoiler needed to neutralize a clickbait. Proc [1]7th Int Work Semant Eval. 2023. p. 1470–5.

26. Rahamim A, Uziel G, Goldbraich E, Tavor AA. Text augmentation using dataset reconstruction for low-resource classification. Find Assoc Comput Linguist ACL 2023. 2023. p. 7389–402.

27. Jiang Y, Qiu R, Zhang Y, Zhang P-F. Balanced and explainable social media analysis for public health with large language models. Australas Database Conf. 2023. p. 73–86.

28. Ilani A, Dolev S. Common Public Knowledge for Enhancing Machine Learning Data Sets. Pro[c] 5th Work Adv tools, Program Lang PLatforms Implement Eval algorithms Distrib Syst. 2023. p. 1–10.

29. Cortes C, Vapnik V. Support-vector networks. Mach Learn. Springer Science and Business Media LLC; 1995;20:273–97.

30. Ye J, Chen X, Xu N, Zu C, Shao Z, Liu S, et al. A comprehensive capability analysis of gpt-3 and GPT3.5 series models. arXiv Prepr arXiv230310420. 2023;

31. OpenAI. GPT4 technical report. 2023.

## Supplementary Material

**Table 2.** S1: The full hyperparameter configurations of the RoBERTa classification model.

| **Hyper-parameter** | **Value** |
|---|---|
| Number of Layers | 12 |
| Hidden size | 768 |
| FFN inner hidden size | 3072 |
| Attention heads | 12 |
| Attention head size | 64 |
| Dropout | 0.1 |
| Attention Dropout | 0.1 |
| Weight Decay | 0.01 |
| Learning Rate Decay | Linear |
| Adam $\beta_1$ | 0.9 |
| Adam $\beta_2$ | 0.98 |
| Gradient Clipping | 0 |
| Batch Size | 32 |
| Warmup Ratio | 0 |
| Adam Epsilon | $1 \times 10^{-8}$ |

**Table 3.** S2: The full prompt for each classification task.

| Task | Prompt |
|---|---|
| Self-report depression | You are a depression detection system based on raw tweet data. The system should analyze the provided tweet and predict whether the user is self-reporting depression or not. Given a tweet as input, the system should output a 1 if the user is self-reporting depression, and 0 otherwise. If a text response is generated, reanalyze the input until a 1 or 0 is generated. |
| Self-report COPD | You are a Chronic Obstructive Pulmonary Disease (COPD) detection system based on raw tweet data. The system should analyze the provided tweet and predict whether the user is self- |
| | reporting COPD or not. Given a tweet as input, the system should output a 1 if the user is self-reporting COPD, and 0 otherwise. If a"text response is generated, reanalyze the input until a 1 or 0 is generated. |
| Self-report breast cancer | You are a breast cancer detection system based on raw tweet data. The system should analyze the provided tweet and predict whether the user is self-reporting breast cancer or not. Given a tweet as input, the system should output a 1 if the user is self-reporting breast cancer, and 0 otherwise. If a text response is generated, reanalyze the input until a 1 or 0 is generated. |
| Change in medications regimen | You are a change in medication regimen detection system based on social media posts. The system should analyze the provided post and predict whether the user is describing a change in medication regimen or not. Given a post as input, the system should output a 1 if the user is describing a change in medication regimen, and 0 otherwise. If a text response is generated, reanalyze the input until a 1 or 0 is generated. |
| Self-report adverse pregnancy outcomes | You are an adverse pregnancy outcome detection system based on social media posts. The system should analyze the provided post and predict whether the user is self-reporting an adverse pregnancy outcome or not. Given a post as input, the system should output a 1 if the user is self-reporting an adverse pregnancy outcome, and 0 otherwise. If a text response is generated, reanalyze the input until a 1 or 0 is generated. |
| Self-report potential cases of COVID-19 | You are an COVID-19 detection system based on social media posts. The system should analyze the provided post and predict whether the user is self-reporting a potential case of COVID-19 or not. Given a post as input, the system should output a 1 if the user is self-reporting a potential case of COVID-19, and 0 otherwise. If a text response is generated, reanalyze the input |
| | until a 1 or 0 is generated. |

**Table 4.** S3: The showcase of data augmentation with GPT4 for self-report depression.

| Source | Post |
|---|---|
| Original | SSRIs have never done much for me’ I've been medication-free for 3 years. I cope well at times, am dysfunctional other times. I hope to be more consistently regulated. I also hope to have fewer intrusive thoughts/memories, and reduce anxiety and depression symptoms. |
| Generated by GPT4 | 3 years off SSRIs and still counting. My journey has highs and lows bu’ I'm striving for balance. H’re's to fewer anxiety episodes and quieter minds. #MentalHealthJourney |
| | Sometimes I handle life well, other times not so much. Been off meds for 3 years. Looking forward to a future where my mind is calmer, less anxious, and less haunted by the past. #HealingJourney |
| | 3 years of being med-free has its share of triumphs and struggles. I yearn for steadier emotions, less anxiety and depression, and a mind free of unwanted thoughts. #MentalHealthAwareness |
| | Three years off meds and still fighting. Some days are easier than others. H’re's to hoping for a future with less anxiety, fewer intrusive thoughts, and more consistency. #MentalWellness |
| | Been off SSRIs for 3 years now. The journey is bumpy bu’ I'm hopeful for more regulated emotions, less anxiety, and a mind free from recurring distressing memories. #MentalHealth |

**Table 5.** S4: The precision, recall, and $F_1$ scores of employing human-annotated data as training data (Human-ann), utilizing LLM-annotated data as training data (GPT3.5-ann and GPT4-ann), and utilizing the LLM as a zero-shot classifier (GPT3.5 zero-shot and GPT4 zero-shot).

| Model | Label source | Precision | Recall | F1 |
|---|---|---|---|---|
| Self-report depression | | | | |
| SVM | Human-ann | 0.68 (±0.01) | 0.70 (±0.07) | 0.69 (±0.04) |
| SVM | GPT3.5-ann | 0.58 (±0.02) | 0.97 (±0.08) | 0.72 (±0.01) |
| SVM | GPT4-ann | 0.58 (±0.02) | 0.99 (±0.02) | 0.73 (±0.01) |
| RoBERTa | Human-ann | 0.76 (±0.05) | 0.88 (±0.07) | **0.82 (±0.02)** |
| RoBERTa | GPT3.5-ann | 0.66 (±0.07) | 0.94 (±0.08) | 0.77 (±0.04) |
| RoBERTa | GPT4-ann | 0.61 (±0.05) | **0.99 (±0.01)** | 0.75 (±0.04) |
| BERTweet | Human-ann | **0.80 (±0.06)** | 0.85 (±0.08) | **0.82 (±0.03)** |
| BERTweet | GPT3.5-ann | 0.70 (±0.06) | 0.88 (±0.10) | 0.78 (±0.02) |
| BERTweet | GPT4-ann | 0.66 (±0.05) | 0.98 (±0.01) | 0.79 (±0.03) |
| SocBERT | Human-ann | 0.78 (±0.03) | 0.85 (±0.07) | 0.81 (±0.04) |
| SocBERT | GPT3.5-ann | 0.68 (±0.05) | 0.88 (±0.05) | 0.77 (±0.02) |
| SocBERT | GPT4-ann | 0.66 (±0.03) | 0.97 (±0.02) | 0.79 (±0.01) |
| GPT3.5 zero-shot | | 0.70 (±0.01) | 0.84 (±0.05) | 0.76 (±0.02) |
| GPT4 zero-shot | | 0.69 (±0.01) | 0.98 (±0.01) | 0.81 (±0.01) |
| Self-report COPD | | | | |
| SVM | Human-ann | 0.65 (±0.02) | 0.65 (±0.03) | 0.65 (±0.01) |
| SVM | GPT3.5-ann | 0.52 (±0.00) | 0.95 (±0.12) | 0.67 (±0.03) |
| SVM | GPT4-ann | 0.52 (±0.00) | 1.00 (±0.00) | 0.68 (±0.00) |
| RoBERTa | Human-ann | 0.84 (±0.04) | 0.91 (±0.05) | **0.87 (±0.03)** |
| RoBERTa | GPT3.5-ann | 0.53 (±0.02) | 0.90 (±0.15) | 0.66 (±0.03) |
| RoBERTa | GPT4-ann | 0.54 (±0.03) | 0.99 (±0.01) | 0.70 (±0.02) |
| BERTweet | Human-ann | **0.87 (±0.05)** | 0.87 (±0.04) | **0.87 (±0.01)** |
| BERTweet | GPT3.5-ann | 0.53 (±0.02) | 0.90 (±0.20) | 0.66 (±0.06) |
| BERTweet | GPT4-ann | 0.56 (±0.03) | 0.97 (±0.03) | 0.71 (±0.02) |
| SocBERT | Human-ann | 0.80 (±0.03) | 0.87 (±0.07) | 0.83 (±0.02) |
| SocBERT | GPT3.5-ann | 0.54 (±0.02) | 0.90 (±0.10) | 0.67 (±0.03) |
| SocBERT | GPT4-ann | 0.54 (±0.01) | **1.00 (±0.01)** | 0.70 (±0.01) |
| GPT3.5 zero-shot | | 0.56 (±0.02) | 0.74 (±0.03) | 0.64 (±0.02) |
| GPT4 zero-shot | | 0.61 (±0.03) | 0.99 (±0.01) | 0.76 (±0.02) |
| Self-report breast cancer | | | | |
| SVM | Human-ann | 0.53 (±0.02) | 0.70 (±0.02) | 0.60 (±0.01) |
| SVM | GPT3.5-ann | 0.36 (±0.01) | 0.74 (±0.03) | 0.48 (±0.02) |
| SVM | GPT4-ann | 0.42 (±0.01) | 0.76 (±0.04) | 0.54 (±0.02) |
| RoBERTa | Human-ann | 0.80 (±0.04) | 0.88 (±0.02) | 0.84 (±0.03) |
| RoBERTa | GPT3.5-ann | 0.38 (±0.01) | 0.88 (±0.06) | 0.53 (±0.01) |
| RoBERTa | GPT4-ann | 0.59 (±0.07) | 0.91 (±0.04) | 0.72 (±0.05) |
| BERTweet | Human-ann | 0.81 (±0.04) | 0.87 (±0.03) | 0.84 (±0.02) |
| BERTweet | GPT3.5-ann | 0.38 (±0.01) | 0.87 (±0.08) | 0.53 (±0.03) |
| BERTweet | GPT4-ann | 0.59 (±0.08) | 0.89 (±0.06) | 0.71 (±0.04) |
| SocBERT | Human-ann | **0.84 (±0.03)** | 0.85 (±0.01) | **0.85 (±0.02)** |
| SocBERT | GPT3.5-ann | 0.38 (±0.03) | 0.85 (±0.06) | 0.52 (±0.02) |
| SocBERT | GPT4-ann | 0.59 (±0.03) | 0.90 (±0.04) | 0.72 (±0.02) |
| GPT3.5 zero-shot | | 0.37 (±0.02) | 0.86 (±0.04) | 0.52 (±0.02) |
| GPT4 zero-shot | | 0.56 | **0.95** | 0.70 |
| | | (±0.01) | **(±0.01)** | (±0.01) |
| Change in medications regimen | | | | |
| SVM | Human-ann | 0.23 (±0.02) | 0.49 (±0.03) | 0.31 (±0.02) |
| SVM | GPT3.5-ann | 0.14 (±0.00) | 0.74 (±0.03) | 0.24 (±0.01) |
| SVM | GPT4-ann | 0.15 (±0.03) | 0.80 (±0.11) | 0.25 (±0.05) |
| RoBERTa | Human-ann | 0.54 (±0.12) | 0.63 (±0.10) | **0.57 (±0.02)** |
| RoBERTa | GPT3.5-ann | 0.19 (±0.02) | 0.79 (±0.11) | 0.31 (±0.03) |
| RoBERTa | GPT4-ann | 0.26 (±0.03) | 0.75 (±0.06) | 0.38 (±0.03) |
| BERTweet | Human-ann | **0.67 (±0.11)** | 0.48 (±0.08) | 0.55 (±0.02) |
| BERTweet | GPT3.5-ann | 0.18 (±0.02) | 0.83 (±0.09) | 0.30 (±0.03) |
| BERTweet | GPT4-ann | 0.26 (±0.03) | 0.73 (±0.10) | 0.38 (±0.03) |
| SocBERT | Human-ann | 0.58 (±0.08) | 0.52 (±0.08) | 0.54 (±0.06) |
| SocBERT | GPT3.5-ann | 0.19 (±0.03) | **0.85 (±0.07)** | 0.31 (±0.03) |
| SocBERT | GPT4-ann | 0.28 (±0.03) | 0.68 (±0.09) | 0.39 (±0.03) |
| GPT3.5 zero-shot | | 0.17 (±0.00) | 0.75 (±0.02) | 0.28 (±0.01) |
| GPT4 zero-shot | | 0.24 (±0.01) | 0.77 (±0.05) | 0.37 (±0.01) |
| Self-report adverse pregnancy outcomes | | | | |
| SVM | Human-ann | 0.70 (±0.01) | 0.67 (±0.01) | 0.68 (±0.01) |
| SVM | GPT3.5-ann | 0.44 (±0.01) | 0.62 (±0.02) | 0.51 (±0.01) |
| SVM | GPT4-ann | 0.45 (±0.01) | 0.78 (±0.02) | 0.57 (±0.01) |
| RoBERTa | Human-ann | 0.82 (±0.04) | **0.92 (±0.03)** | 0.87 (±0.01) |
| RoBERTa | GPT3.5-ann | 0.46 (±0.02) | 0.65 (±0.05) | 0.54 (±0.01) |
| RoBERTa | GPT4-ann | 0.49 (±0.02) | 0.77 (±0.06) | 0.60 (±0.02) |
| BERTweet | Human-ann | **0.88 (±0.03)** | 0.89 (±0.03) | **0.88 (±0.01)** |
| BERTweet | GPT3.5-ann | 0.46 (±0.03) | 0.65 (±0.04) | 0.54 (±0.02) |
| BERTweet | GPT4-ann | 0.49 (±0.01) | 0.75 (±0.04) | 0.59 (±0.02) |
| SocBERT | Human-ann | 0.85 (±0.03) | 0.87 (±0.03) | 0.86 (±0.01) |
| SocBERT | GPT3.5-ann | 0.46 (±0.01) | 0.65 (±0.05) | 0.54 (±0.02) |
| SocBERT | GPT4-ann | 0.48 (±0.02) | 0.77 (±0.02) | 0.59 (±0.02) |
| GPT3.5 zero-shot | | 0.49 (±0.01) | 0.62 (±0.02) | 0.55 (±0.01) |
| GPT4 zero-shot | | 0.52 (±0.01) | 0.81 (±0.01) | 0.63 (±0.01) |
| Self-report potential cases of COVID-19 | | | | |
| SVM | Human-ann | 0.59 (±0.05) | 0.28 (±0.03) | 0.38 (±0.02) |
| SVM | GPT3.5-ann | 0.21 (±0.01) | 0.71 (±0.03) | 0.33 (±0.01) |
| SVM | GPT4-ann | 0.43 (±0.02) | 0.38 (±0.03) | 0.40 (±0.02) |
| RoBERTa | Human-ann | **0.68 (±0.07)** | 0.64 (±0.11) | 0.66 (±0.06) |
| RoBERTa | GPT3.5-ann | 0.23 (±0.03) | **0.86 (±0.07)** | 0.36 (±0.03) |
| RoBERTa | GPT4-ann | 0.48 (±0.06) | 0.60 (±0.09) | 0.52 (±0.03) |
| BERTweet | Human-ann | **0.68 (±0.06)** | 0.71 (±0.07) | **0.69 (±0.02)** |
| BERTweet | GPT3.5-ann | 0.23 (±0.03) | 0.79 (±0.13) | 0.35 (±0.04) |
| BERTweet | GPT4-ann | 0.43 (±0.03) | 0.61 (±0.02) | 0.51 (±0.01) |
| SocBERT | Human-ann | 0.66 (±0.06) | 0.67 (±0.06) | 0.66 (±0.01) |
| SocBERT | GPT3.5-ann | 0.24 (±0.01) | 0.83 (±0.05) | 0.38 (±0.01) |
| SocBERT | GPT4-ann | 0.49 (±0.02) | 0.57 (±0.04) | 0.53 (±0.02) |
| GPT3.5 zero-shot | | 0.23 (±0.01) | 0.74 (±0.03) | 0.35 (±0.01) |
| GPT4 zero-shot | | 0.45 | 0.65 | 0.53 |

| | | | |
|---|---|---|---|
| | (±0.02) | (±0.03) | (±0.02) |